\documentclass{ecai2012}
\usepackage{mathrsfs}
\usepackage{amssymb}
\usepackage{graphicx}
\usepackage{subfigure}
\usepackage{amsmath}
\usepackage{graphicx}
\usepackage{dcolumn}
\usepackage{bm}

\newcommand{\be}{\begin{equation}}
\newcommand{\ee}{\end{equation}}
\newcommand{\bea}{\begin{eqnarray}}
\newcommand{\eea}{\end{eqnarray}}

\newcommand{\ie}{\textit{i.e.~}}

\newcommand{\prs}[1]{{\left(#1\right)}}
\newcommand{\prob}[1]{{\mathcal{P}\prs{#1}}}

\begin{document}

\title{Comparative Study for Inference of Hidden Classes in Stochastic Block Models}
\author{Pan Zhang\institute{CNRS and ESPCI
    ParisTech, 10 rue Vauquelin, UMR 7083 Gulliver, Paris
    75005,France, email: pan.zhang@espci.fr} \and Florent Krzakala\institute{CNRS and ESPCI
    ParisTech, 10 rue Vauquelin, UMR 7083 Gulliver, Paris 75005,
    France, email: fk@espci.fr} \and J\"org Reichardt\institute{Institute for Theoretical Physics, University of W\"urzburg Am Hubland 97074 W\"urzburg, Germany, email:reichardt@physik.uni-wuerzburg.de} \and Lenka
  Zdeborov\'a\institute{Institut de Physique Th\'eorique, IPhT, CEA
    Saclay, and URA 2306, CNRS, 91191 Gif-sur-Yvette, France, email:
    lenka.zdeborova@gmail.com}}
\maketitle
\bibliographystyle{ecai2012}

\begin{abstract} 
Inference of hidden classes in  stochastic block model is a classical problem with important applications. Most commonly used methods for this problem involve na\"{\i}ve mean field approaches or heuristic spectral methods. Recently, belief propagation was proposed for this problem. In this contribution we perform a comparative study between the three methods on synthetically created networks. We show that belief propagation shows much better performance when compared to na\"{\i}ve mean field and spectral approaches. This applies to accuracy, computational efficiency and the tendency to overfit the data. 
\end{abstract}

\section{Introduction}
A large portion of the intriguing emergent phenomena of complex many particle systems is a consequence of the structure of interactions among their constituents. Bluntly, a soup of neurons does not have the same capabilities as a specifically woven neural net. Similar considerations apply to social systems, information systems, biological systems or economical systems where the patterns of interaction are far from random and result in complex system-wide phenomena. 

Fueled by a flood of readily available relational data, recent years have seen a surge of research focused on structural properties of networks as first step to understanding some of the properties of complex systems and ultimately their function \cite{EasleyKleinberg,NewmanBook}.   

Interestingly, it is often much easier to map the network of interactions than to explain its function. A prime example of this phenomenon are protein interaction networks. Modern biotechnology allows to automatize charting the matrix of pairwise binding relations for all proteins produced by an organism, \ie do two proteins form a stable link or not \cite{Uetz2000}. As proteins generally operate in complexes (agglomerates of several proteins) such a network of pairwise interactions encodes latent information about protein function. Hence, it makes sense to use network structure to make inferences about protein function or plan and guide other wet-lab experiments aimed at elucidating function \cite{Pinkert2010}. Similar considerations apply to the analysis of social networks where interactions are recorded in online data streams but information on the properties of the actual agents remains hidden behind pseudonyms or avatars \cite{Wellman2001}. 

Hence, the hypothesis behind network analysis is that nodes in a network which have similar patterns of interaction are likely to have common properties or perform similar function. Discovering topological similarities and differences thus hints at the existence of possible latent features of the nodes in the network that merit further analysis. 

Being a first step to more detailed analysis, such exploratory analysis is often highly consequential. It is important to thoroughly understand the algorithms used in every detail and to be aware of possible limitations and pitfalls. This contribution aims at raising this awareness using the simple example of inferring the parameters of a Poisson-mixture model, the so-called Stochastic-Block-Model (SMB) \cite{HollandLaskeyLeinhard1983,Goldenberg2009}, in undirected unweighted unipartite networks. The conclusions we draw, however, extend well beyond this example and we discuss these consequences at the end of the paper. 

Our contribution is then organized as follows: first we introduce the stochastic block model as a way to capture density fluctuations in relational datasets and infer latent variables. Next, we discuss the theoretical limitations that any inference technique for such a model must face: namely a sharp transition between a parameter region where inference is feasible and a parameter region where inference is impossible. Third, we briefly review spectral approaches and the Expectation Maximization (EM) algorithm in conjunction with the na\"{\i}ve mean field approach. We then introduce a formulation of the EM algorithm based on belief propagation. Fourth, we compare the performance of these three approaches on ensembles of benchmark networks from a region near the above mentioned feasible-infeasible transition in the parameter space. In this region, particularly difficult problem instances can be found that allow to highlight performance differences.  Finally, we discuss our findings and the possible extensions and consequences to other models and inference tasks. 

\section{The Stochastic Block Model}\label{sec:SBM}
The simplest model of a network of $N$ nodes and $M$ undirected unweighted edges between them is a an Erd\H{o}s-R\'enyi graph. It assumes that a link falls between any pair of nodes $(i,j)$ with constant probability $p_{ij}=p_0=2M/[N(N-1)]$, independently of whether links exist between other pairs of nodes. Consequentially, large networks with low link density $p_0$ generated by this model have a Poissonian degree distribution with mean degree $\langle k\rangle=p_0(N-1)$. This model can already explain two main characteristics of real world networks - their small world property of short average path lengths and their connectedness even at low densities. Unfortunately, it cannot explain much more. In particular, it fails to capture the large variance of link densities between groups of nodes observed in many networks. 

In real world networks, not all nodes are created equal and may represent entities of very different properties or functions. Whether two nodes are linked often depends on these properties. Consider the example of protein interaction again. Membrane proteins will certainly bind to other membrane proteins to form stable membranes, but, for example, the enzymes involved in various catalytic reactions should not stick to the cell membrane since otherwise the interior of the cell would soon be depleted of these essential molecules \cite{Pinkert2010}. In an entirely different social context, one will certainly observe social interactions correlated with the agents' age, gender, possibly income or education. Social ties will depend on these qualities an thus network structure is indicative of node properties and may be used to make corresponding inferences.

One of the simplest models capable of capturing the dependence of link probability on node properties is the Stochastic Block Model \cite{HollandLaskeyLeinhard1983}. It assumes that each node $i\in\{1,..,N\}$ is of one and only one of $q$ classes and $t_i=r$ is indicating the membership of node $i$ in class $r\in\{1,..,q\}$. As before, nodes are linked independently, but now, the probability of node $i$ linking to node $j$ depends on $t_i$ and $t_j$ alone, \ie $p_{ij}=p_{t_it_j}$. One can easily write down a probabilistic generative model for this sort of network. First, we assume that nodes are assigned into $q$ classes randomly by a multinomial distribution with parameters $\prob{t_i=r}=p_r$. Next, we specify the matrix of link probabilities between classes $p_{rs}\in{(0,1)^{q\times q}}$. Our set of parameter thus comprises of $\theta=\{q,{p_r},{p_{rs}}\}$. The probability of generating a specific $\{0,1\}^{N\times N}$ adjacency matrix $\mathbf{A}$ together with a specific assignment of nodes into classes $\mathbf{t}$ is then given as:
\begin{equation}
\label{CDLikelihood}
\prob{\mathbf{A},\mathbf{t}|\theta} = \prod_{i<j}\left[p_{t_it_j}^{A_{ij}}(1-p_{t_it_j})^{(1-A_{ij})}\right]\prod_i p_{t_i}
\end{equation}
The expected average density of links in such a network is $p_0=\sum_{rs}p_rp_{rs}p_s$.
If we were able to observe the adjacency  matrix $\mathbf{A}$ \emph{and} class memberships $\mathbf{t}$ at unknown parameters, equation (\ref{CDLikelihood}) would give us the complete data likelihood of the parameters $\theta$. It is then easy to estimate the parameters $\theta^*$ which maximize (\ref{CDLikelihood}):
\begin{eqnarray}
\label{ParamterEstimation}
p_r = &\frac{1}{N}\sum_i\delta_{t_i,r}\\
p_{rs} = &\frac{1+\delta_{rs}}{N p_r (N p_s-\delta_{rs})}\sum_{i<j}A_{ij}\delta_{t_i,r}\delta_{t_j,s}\nonumber
\end{eqnarray}
With (\ref{CDLikelihood}) being a member of the exponential family,  these estimators are consistent, efficient and the model is identifiable, \ie the maxima are unique. In this contribution we always assume that the correct number of classes $q$ is known. 

However, in practical applications as discussed, the situation is often that we only have access to the adjacency matrix $\mathbf{A}$ but \emph{not} to the class labels $\mathbf{t}$ which are our primary interest for explaining network structure and possibly function. Fortunately, under certain circumstances we can still draw conclusions about these hidden variables using the toolbox of statistical inference. What these circumstances are and how this is usually done will be discussed in the following two sections. 

\section{General Considerations}\label{sec:GC}
It is clear that the task of inferring the unobserved latent variables is only possible if the preference matrix $p_{rs}$ shows sufficient ''contrast''. If all entries were the same, \ie $p_{rs}=p_0$, then of course no method can perform any inference on the hidden variables. Conversely, if $p_{rs}=p_0\delta_{r,s}$, then the network practically consists of several disconnected components and inference reduces to the trivial task of identifying the component to which an individual node belongs. Between these two extremes of impossible and trivial, there is a sharp phase transition \cite{Reichardt2008,PhysRevLett.107.065701,PhysRevE.84.066106}. It divides the parameter space into a region where it is provably impossible to infer the latent variables with an accuracy higher than guessing and a region where it is possible with high accuracy. 
  
Theoretical analysis has shown that the transition persists in infinitely large networks when they are sparse, \ie the average degree per node does not grow with the system size. In other words, networks in which the elements of $p_{rs}$ scale as $1/N$. In contrast, for dense networks in which $p_{rs}$ does not scale with $N$, considering larger networks means considering proportionally larger average degrees and this will render even very small amounts of variance in $p_{rs}$ detectable and thus lets the region of impossible inference vanish \cite{Onsjo2006}. 

In real applications, we cannot generally increase network size at constant parameters. We will observe both the region of impossible and possible inference. However, the parameter region of impossible inference will be smaller for denser networks, \ie those with higher average degree. Further, it has been shown that networks with parameters in the vicinity of the transition point are the instances in which inference is hardest \cite{PhysRevLett.107.065701,PhysRevE.84.066106}. 

As it is our aim to highlight performance differences between different inference techniques for the SBM, we will focus our attention on instances in sparse graphs near the transition from impossible to possible inference. Before we come to this analysis, however, we will introduce the contestants. 

\section{Inferring Stochastic Block models}\label{sec:SBMInf}
When inferring latent structure in data, one can take the route of statistical inference if one can justify a statistical model to fit to the data as we've done with the SBM. It may also be sensible to use a simple dimensionality reducing heuristic. We consider both of these approaches.

\subsection{Spectral Approaches}
When dealing with high dimensional data such as networks and searching for common patterns of interactions, a natural strategy is to try reducing the dimensionality in such a way that nodes with similar interaction partners are mapped to positions in some low dimensional space, while nodes with very different interaction partners should be positioned far apart. One then uses standard clustering algorithms, such as $k$-means in our case, originally developed for multivariate data and to analyze the nodes in their low dimensional embedding. This is the strategy behind all spectral techniques of network analysis.

Let us consider the adjacency matrix $\mathbf{A}$ as a list of $N$ measurements in an $N$-dimensional feature space, each row describing one node in $N$ dimensions, namely, its relations to the other nodes. We could then apply a variant of multidimensional scaling such as principal component analysis (PCA). We would subtract the means of the measurements in each dimension, calculate the covariance matrix and find the directions of maximum variance by an eigen-decomposition of the co-variance matrix. Finally, we would project our data matrix onto the first $q$ principal components, \ie those eigenvectors of the covariance matrix corresponding to the largest eigenvalues. 

A method similar in spirit has been introduced specifically for networks \cite{PhysRevE.74.036104}. It differs from PCA only slightly in that it not only removes the means of the rows, but, since $\mathbf{A}$ is symmetric, also the means of the columns. This is to say, the original matrix $\mathbf{A}$ is transformed into a so called modularity matrix $\mathbf{B}$  via
\begin{equation}
B_{ij} = A_{ij}-\frac{k_ik_j}{2M}.
\end{equation}
This modularity matrix $\mathbf{B}$ now has row-sums and column-sums zero. Note that the terms $k_ik_j/2M\ll 1$ for sparse networks. Since $\mathbf{B}$ is symmetric, the eigenvectors of a corresponding ``covariance matrix'' $\mathbf{C}=\mathbf{B}\mathbf{B}^{\mathrm{T}}$ are the eigenvectors of $\mathbf{B}$ and hence the projection of the modularity matrix onto the ``principal components'' is given directly by the components of the eigenvectors corresponding to the largest magnitude eigenvectors of $\mathbf{B}$. This approach has recently been claimed to be no worse than any other approach \cite{PhysRevLett.108.188701} and we will evaluate this claim in this paper.

Another aspect of this method is worth mentioning. It is known that the best rank-$q$ approximation to a symmetric matrix is given by its eigen-decomposition retaining only the $q$ eigenvalues largest in magnitude. ``Best'' here means in terms of reconstruction error under the Frobenius norm. If $\mathbf{V}$ is a matrix the columns of which are the eigenvectors of $\mathbf{B}$ ordered by decreasing magnitude of the corresponding eigenvalue, then the entries of the optimal rank-$q$ approximation $\mathbf{B}'$ will be given by
\begin{equation}
B'_{ij}=\sum_{r=1}^q V_{ir}\lambda_rV_{jr}.
\end{equation}
So we see that $B_{ij}'$ is large when the rows $i$ and $j$ of $\mathbf{V}$ are parallel \emph{and} all the considered $\lambda_r$ with $r\in\{1,..,q\}$ are positive. In contrast, if all $\lambda_r$ are negative,  rows $i$ and $j$ of  $\mathbf{V}$ should be anti-parallel to make $B_{ij}'$ large. Large positive eigenvalues are indicative of block models with some $p_{rr}$ large while large negative eigenvalues are indicative of block models with some $p_{rr}$ small in comparison to the average density of the network $p_0$. We can conclude that when these cases mix, it will generally be very difficult to find an embedding that maps nodes from a network with similar interaction patterns to positions that are close in space using spectral decomposition of the modularity matrix.

Instead of using an embedding that minimizes a reconstruction error, one can also introduce a pairwise similarity measure based on the network topology and then find an embedding of the $N\times N$ similarity matrix such that ``similar nodes'' are ``close''. This approach is implemented in the widely used diffusion-map \cite{Lafon2006}.

Assume a random walker is traversing the network. When at node $i$, the walker will then move to any node $j\neq i$  with probability $p_{j|i}=A_{ij}/k_i$. Here, $k_i=\sum_j A_{ij}$ is the number of neighbors of node $i$. We can identify in $p_{j|i}$ as the entries of an $N\times N$ row stochastic transition matrix $\mathbf{P}=\mathbf{D}^{-1}\mathbf{A}$ where $\mathbf{D}$ is a diagonal matrix with $D_{ii}=k_i$.
The probability that the random walker, after starting in node $i$, reaches node $j$ in exactly $t$ steps is then given as $p_t(j|i)\equiv \mathbf{P}^t_{ij}$. The stationary distribution of the random walker on the $N$ nodes of the network is given by $\pi_0^i\equiv\lim_{t\to\infty}p_t(i|j)=k_i/2M$. Equipped with these definitions, one can define a ''diffusion distance'' between nodes $i$ and $j$ via
\begin{equation}
D^2_t(i,j)=\sum_k \frac{(p_t(k|i)-p_t(k|j))^2}{\pi_0^k}.
\end{equation}
This is a sensible measure of topological distance between nodes $i$ and $j$ as it measures a difference in the distributions of arrival sites when the random walker starts from either $i$ or $j$. One can find an optimal embedding such that the Euclidean distance in the low dimensional space matches the diffusion distance to any desired precision. The coordinates of this embedding are given by the entries in the eigenvectors corresponding to the $q$ largest non-trivial right eigenvectors of $\mathbf{P}$ scaled by the corresponding eigenvalue to power $t$. Since the largest right eigenvalue of $\mathbf{P}$ is always $\lambda_1=1$ and the corresponding eigenvector is constant, it is considered trivial. If a match to relative precision $\delta$ is required we must include all eigenvectors $\mathbf{v}_r$ of $\mathbf{P}$ with $|\lambda_r|^t>\delta |\lambda_2|^t$ where the $\lambda$ are the right eigenvalues of $\mathbf{P}$. As all eigenvalues of $\mathbf{P}$ are smaller in magnitude than $1$, $\lambda_2$ dominates for large $t$ and thus the large scale structural features.  
In this case, large negative eigenvalues are not a problem, since the embedding is such that Euclidian distance between the positions of the nodes in the low dimensional space approximates the topological distance and not the scalar product dressed with the eigenvalues as in the case of the spectral decomposition.

\subsection{Expectation Maximization}
The goal of maximum likelihood inference aims to estimate parameters for a generative model such that the observed data becomes maximally likely under this model. Our generative model (\ref{CDLikelihood}) gives us the probability of observing the network \emph{and} the node classes. If only the network is observed we need to trace out the node classes. Specifically, we seek
\begin{equation}
\theta^*=\mathrm{argmax}_\theta \mathcal{L}(\theta)\equiv\log\sum_{\mathbf{t}}\prob{\mathbf{A},\mathbf{t}|\theta}.
\end{equation}
The sum over all possible assignments of nodes into latent classes is computationally intractable and so one resorts defining a lower bound on the log-likelihood $\mathcal{L}(\theta)$ which can be both  evaluated and maximized. This bound is know as the Free Energy 
\begin{equation}
\mathcal{F}(\tilde{\mathcal{P}}(\mathbf{t})
,\theta)\equiv \sum_{\boldmath{t}}\tilde{\mathcal{P}}(\mathbf{t})\log \prob{\mathbf{A},\mathbf{t}|\theta} - \sum_{\boldmath{t}}\tilde{\mathcal{P}}(\mathbf{t})\log \tilde{\mathcal{P}}(\mathbf{t}).
\label{FreeEnergy}
\end{equation}
The Free energy $\mathcal{F}$ is a functional of a distribution over the latent variables $\tilde{\mathcal{P}}(\mathbf{t})$ and the model parameters $\theta$. It is easily shown that $\mathcal{F}$ is indeed a lower bound on $\mathcal{L}(\theta)$:
\begin{equation}
\mathcal{F}(\tilde{\mathcal{P}}(\mathbf{t}),\theta) = -D_{\rm{KL}}(\tilde{\mathcal{P}}(\mathbf{t})||\prob{\mathbf{t}|\mathbf{A,\theta}}) + \mathcal{L}(\theta).
\end{equation}
and that if $\mathcal{F}$ has a (global) maximum in $(\tilde{\mathcal{P}}^*(\mathbf{t}),\theta^*)$ then $\mathcal{L}(\theta)$ also has a (global) maximum in $\theta^*$ \cite{Neal98aview}. The procedure for maximizing $\mathcal{F}$ in turn with respect to its two arguments is known as the Expectation Maximization algorithm \cite{Dempster1977}. Specifically, maximizing $\mathcal{F}$ with respect to $\tilde{\mathcal{P}}(\mathbf{t})$ at fixed $\theta$ is known as the ''E-Step'', while maximizing $\mathcal{F}$ with respect to $\theta$ at fixed $\tilde{\mathcal{P}}(\mathbf{t})$ is known as the ''M-Step''. Ideally, the E-step tightens the bound by setting $\tilde{\mathcal{P}}(\mathbf{t}) = \prob{\mathbf{t}|\mathbf{A,\theta}}$, but for our model (\ref{CDLikelihood}) the calculation of $\prob{\mathbf{t}|\mathbf{A,\theta}}$ is also intractable. Note that this is in contrast to estimating the parameters of a mixture of Gaussians where, for observed data $\mathbf{X}$, we can easily evaluate $\prob{\mathbf{t}|\mathbf{X,\theta}}$.

Two routes of approximation now lie ahead of us: the first one is to restrict ourselves to a simple factorizable form of $\tilde{\mathcal{P}}(\mathbf{t})=\prod_i\tilde{\mathcal{P}}(t_i)$ which leads to the mean field approach. The second route 
leads to belief propagation.
%
\subsection{E-Step and M-Steps using the na\"{\i}ve mean field}
We shall start by the mean field equations as used for the SBM for instance in \cite{Daudin_2008} or \cite{HofmanWiggins08}. In addition to the assumption of a factorizing $\tilde{\mathcal{P}}(\mathbf{t})$, one introduces the following shorthand: $\psi^i_r \equiv  \tilde{\mathcal{P}}(t_i=r)$. Then, the free energy in the na\"{\i}ve mean field approximation is given by
\begin{eqnarray}
\mathcal{F}_{\rm MF}= & \sum_{i<j,rs} \left(A_{ij}\log\frac{p_{rs}}{1-p_{rs}} + \log(1-p_{rs})\right)\psi^i_r\psi^j_s\nonumber\\
& + \sum_{i,r}\psi^i_r(\log p_r - \log \psi^i_r)
\end{eqnarray}
This free energy is to be maximized with respect to the $\psi^i_r$ by setting the corresponding derivatives to zero and we obtain a set of self-consistent equations the $\psi^i_r$ have to satisfy at $\nabla_\psi \mathcal{F}=0$:
\begin{eqnarray}
\label{MFEstep}
\psi^i_r = & \frac{p_re^{h^i_r}}{\sum_{s}p_{s}e^{h^i_{s}}}\\
h_r^i=&\sum_{j\neq i,s}A_{ij}\log\frac{p_{rs}}{1-p_{rs}}\psi_s^j + \sum_s(N-\delta_{rs}) p_s\log (1-p_{rs})\nonumber
\end{eqnarray} 
The beauty of this approach is its apparent computational simplicity as an update of $\tilde{\mathcal{P}}(\mathbf{t})$ can be carried out in $\mathcal{O}(N\langle k\rangle q^2)$ steps. 
Setting $\nabla_\theta \mathcal{F}_{\rm MF}$ equal to zero and observing  the constraint that $\sum_rp_r=1$,  we derive the following equations for the M-step:
\begin{eqnarray}
\label{MFMstep}
p_r = &\frac{1}{N}\sum_i \psi_r^i\\
p_{rs}= &\frac{\sum_{i<j}A_{ij}\psi^i_r\psi^j_s}{\sum_{i<j}\psi^i_r\psi^j_s}\nonumber
\end{eqnarray}
Note the similarities between eqns.\ (\ref{MFMstep}) and (\ref{ParamterEstimation}).

\subsection{E-Step and M-Steps using Belief Propagation}

Belief propagation equations for mixture models were used by several authors, see e.g. \cite{Hastings06,Getoor,ReichardtSaad2011}. Several important nuances in the algorithm make us adopt belief propagation algorithm for SBM as developed in \cite{PhysRevLett.107.065701,PhysRevE.84.066106}, the implementation can be dowloaded at http://mode\_net.krzakala.org/.

There are several ways one can derive the Belief Propagation equations (see for instance \cite{Yedidia_etal_TR2001-22}). 
One way is from a recursive computation of the free energy under the assumption that the graphical model is a tree.  Application of the same equations on loopy graphical models is then often justified by the fact that correlations between variables induced by loops decay very fast and are hence negligible in the thermodynamic limit. In the case treated here, even when the adjacency graph $A_{ij}$ is sparse, the graphical model representing the probability distribution (\ref{CDLikelihood}) is a fully connected graph on $N$ nodes. However, for sparse networks the interaction for nodes that are not connected by an edge is weak $1-p_{rs}\approx 1$ and the network of strong interactions is locally tree-like. This puts us in the favorable situation of decaying correlations. This was used in \cite{PhysRevLett.107.065701,PhysRevE.84.066106} to argue heuristically that in the limit of large $N$ the belief propagation approach estimates asymptotically exact values of the marginal probabilities $\psi^i_r$ and of the log-likelihood, in a special case of block model parameters this has been proven rigorously in \cite{Sly2012}.

To write the belief propagation equations for the likelihood (\ref{CDLikelihood}) we define conditional marginal probabilities, or \emph{messages}, denoted $\psi_{r}^{i\to j}\equiv \prob{t_i=r|\mathbf{A}\backslash A_{ij},\theta}$. 
This is the marginal probability that the node $i$ belongs to group $r$ in the absence of node $j$.  In the tree approximation we then assume that the only correlations between $i$'s neighbors are mediated through $i$, so that if $i$ were missing---or if its group assignment was fixed---the distribution of its neighbors' states would be a product distribution.  In that case, we can compute the message that $i$ sends $j$ recursively in terms of the messages that $i$ receives from its other neighbors $k$ \cite{PhysRevLett.107.065701,PhysRevE.84.066106}:
\begin{eqnarray}
\psi^{i\to j}_r = & \frac{p_re^{h^{i\to j}_r}}{\sum_{s}p_{s}e^{h^{i\to j}_{s}}}\\
h_r^{i\to j}=&\sum_{k\neq i,j}  \log{  \left[ \sum_s \left( \frac{p_{rs}}{1-p_{rs}}  \right)^{A_{ik}} (1-p_{rs})  \psi_s^{k\to i} \right]  } 
\label{BPEstep}
\end{eqnarray} 
The marginal probability $\psi_r^i$ is then recovered from the messages using (\ref{MFEstep}) and 
\be
       h_r^i = \sum_{j\neq i} \log{\left[  \sum_s   \left( \frac{p_{rs}}{1-p_{rs}}  \right)^{A_{ij}} (1-p_{rs})  \psi_s^{j\to i}   \right]}\, .
\ee
Compared with equations (\ref{MFEstep}), updating the belief propagation equations takes $\mathcal{O}(N^2q^2)$ steps. 

Most real world networks, however, are relatively sparse, \ie the number of edges is much smaller than $N^2$. For such cases the BP equations can be simplified. To see this we consider $c_{rs}=N p_{rs}=O(1)$, in the limit $N\to \infty$ terms $o(N)$ can be neglected as in \cite{PhysRevE.84.066106}, one then needs to keep and update messages $\psi_r^{i\to j}$ only when $A_{ij}=1$. The update equation for field $h_r^{i\to j}$ then is
\be
   h_r^{i\to j}=\sum_{k\in \partial i \setminus j }  \log{  \left( \sum_s c_{rs} \psi_s^{k\to i} \right)  }  - \frac{1}{N} \sum_{k=1}^N \sum_s c_{rs} \psi_s^{k} \, ,\label{BP_iter} 
\ee
where $\partial i$ denotes $i$'s neighborhood.
In order to get the marginal probability $\psi_r^i$  one uses eq.~(\ref{MFEstep})  and 
\be
  h_r^{i}=\sum_{k\in \partial i }  \log{  \left( \sum_s c_{rs} \psi_s^{k\to i} \right)  }  - \frac{1}{N} \sum_{k=1}^N \sum_s c_{rs} \psi_s^{k} \, . 
\ee
Note that it is possible to implement the update of all fields $h_r^{i}$ in $\mathcal{O}(N\langle k\rangle q^2)$ steps, thus making the BP approach as fast the the na\"{\i}ve mean field. In order to do that, we compute the second term in eq.~(\ref{BP_iter}) once at the beginning and then we only add and subtract the contributions to this term that changed.

Once the fixed point of BP equations is found, one uses the Bethe formula to compute the free energy \cite{Yedidia_etal_TR2001-22}
\be
    \mathcal{F}_{\rm BP} =  \frac{1}{N}\sum_{(ij) \in E}  \log{Z^{ij}} - \frac{1}{N} \sum_i \log{\left(\sum_s p_s e^{h^i_s}\right)} - \frac{\langle k\rangle}{2} \label{Bethe_fe}\, ,
\ee
where 
\be
     Z^{ij} = \sum_{r,s} c_{rs} \psi^{i\to j}_r \psi_s^{j\to i} \nonumber
\ee
Again the Bethe free energy is exact if the graphical model is a tree and is a very good approximation to the true free energy in many practical cases, and often a much better one than the MF free energy. An important point is that the Bethe free energy is not guarantied to be a bound on the log-likelihood.

Setting $\nabla_\theta \mathcal{F}_{\rm BP}$ equal to zero and observing that the BP equations are stationarity conditions for the Bethe free energy, one derives the following equations for the M-step of expectation maximization
\bea
        p_r &=& \frac{1}{N}\sum_i \psi^i_r\, , \label{Nish_na_BP}\\
        c_{rs} &=& \frac{1}{N}  \frac{1}{p_r p_s}  \sum_{(i,j) \in E} \frac{c_{rs} ( \psi_r^{i\to j} \psi_s^{j\to i}+ \psi_s^{i\to j} \psi_r^{j\to i} )}{Z^{ij}}.\nonumber
\eea

\section{Performance Comparison}\label{sec:PC}
%
%

\begin{figure*}[ht]
  \center{
    \includegraphics[width= 0.7\linewidth]{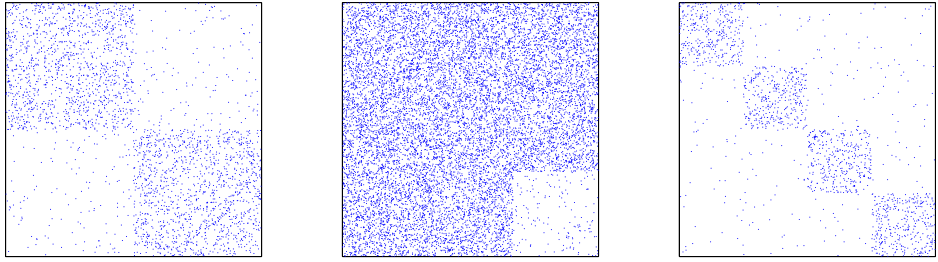}
  }
  \caption{
Adjacency matrices representing the block
  structure used for generating the various examples of the block model eq.~(\ref{CDLikelihood}) in this contribution.  Rows and columns are ordered such that rows/columns corresponding to nodes with the same $t_i$ are next to each other. From left to right: a $q=2$ modular network, a core-periphery structure, and a $q=4$ modular network. }
  \label{fig_bloc} 
\end{figure*}

We will compare the performance of the three approaches presented in the last section on ensembles of test networks which have been generated from (\ref{CDLikelihood}). Hence, we know the true assignment of nodes into classes $t_i$ for all nodes $i\in \{1,..,N\}$. Let us denote by $t^*_i$ the estimates of group assignment that follow from the above algorithms. A simple performance measure is then the ``overlap'' between $\{t_i\}$ and $\{t_i^*\}$ defined as 
\begin{equation} 
Q \equiv \frac{1}{N} \max_{\pi} \sum_i \delta(t^*_i,\pi(t_i)). 
\label{over_true} 
\end{equation} 
Since the $t_i$ can only be recovered up to permutation of the class labels, the maximum over all possible permutations of $\pi$ on $q$ elements is taken. Note that a trivial estimate would be $t_i^*=\mathrm{argmax}_r p_r\,  \forall i$. Hence, only values of $Q>\max_r p_r$ should be considered as successful inference.

\subsection{Belief Propagation vs Mean Field}

To make a comparison of BP and MF we will assume in both approaches that the parameters $p_r$, $p_{rs}$, and the right number of groups $q$ are known. Both approaches output the estimates of marginal probabilities $\psi^i_r$.  In order to estimate the original group assignment, we assign to each node its most-likely group, i.e.
 \be t_i^* = \mathrm{argmax}_r \psi^i_r \, . \label{marginalization} \ee
 If the maximum of $\psi^i_r$ is not unique, we choose at random from all the $q_i$ achieving the maximum.  We refer to this method of estimating the groups as \emph{marginalization}.  Indeed, a standard result show that it is the optimal estimator of the original group assignment $\{t_i\}$ if we seek to maximize the number of nodes at which $t_i=t^*_i$. 

In practical situations, when the true assignment is not known, one can also use the estimates of the marginal probabilities $\psi^i_r$ to compute the confidence of the method about the estimate $t_i^*$ defined as
\be
C \equiv  \frac{1}{N} \sum_i \psi^i_{t^*_i}  \, . \label{over_conf}
\ee 
An important remark is that if the marginals $\psi^i_r$ were evaluated exactly then in the large $N$ limit the overlap and confidence quantities agree, $C=Q$. In our tests the quantity $C-Q$ hence measures the amount of illusive confidence of the method. Values of $C-Q$ larger than zero are very undesirable as they indicate a misleading correlation, and give an illusive information on the amount of information reconstructed. 

%
\begin{figure*}[!ht]
\includegraphics[width=0.32\linewidth]{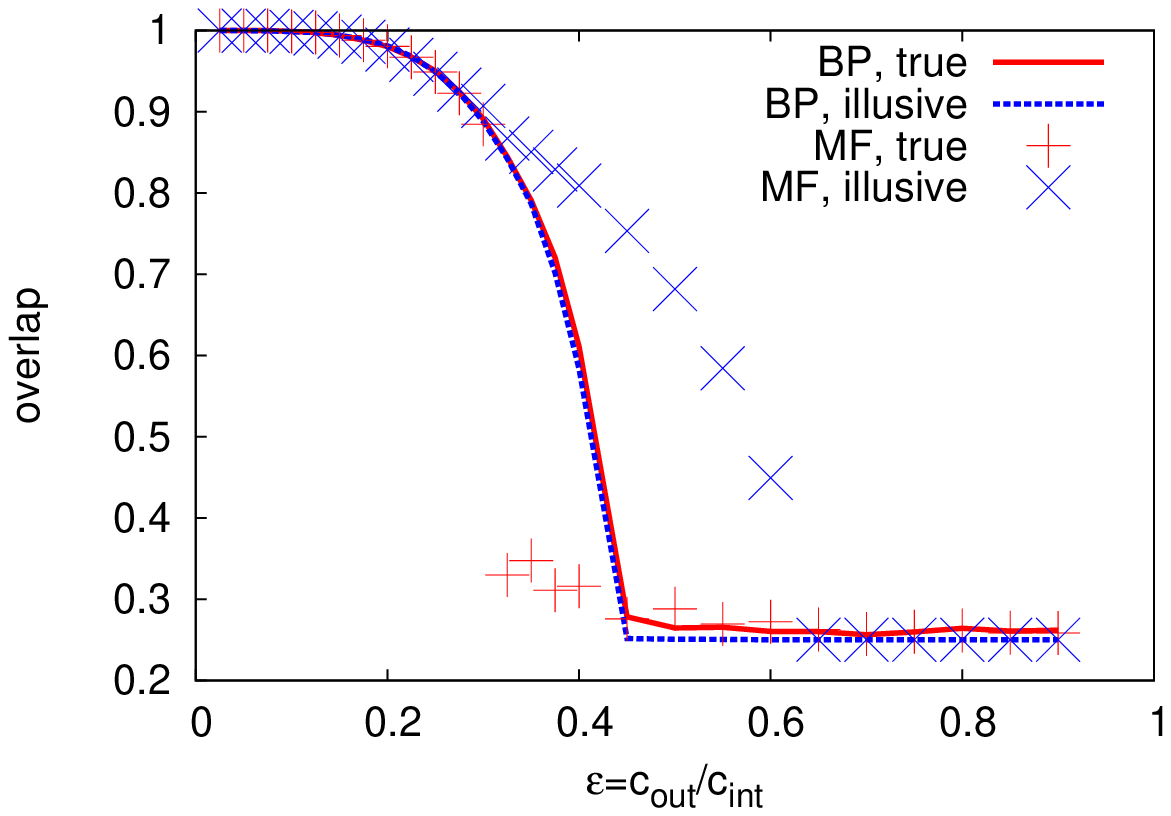} 
  \includegraphics[width=0.32\linewidth]{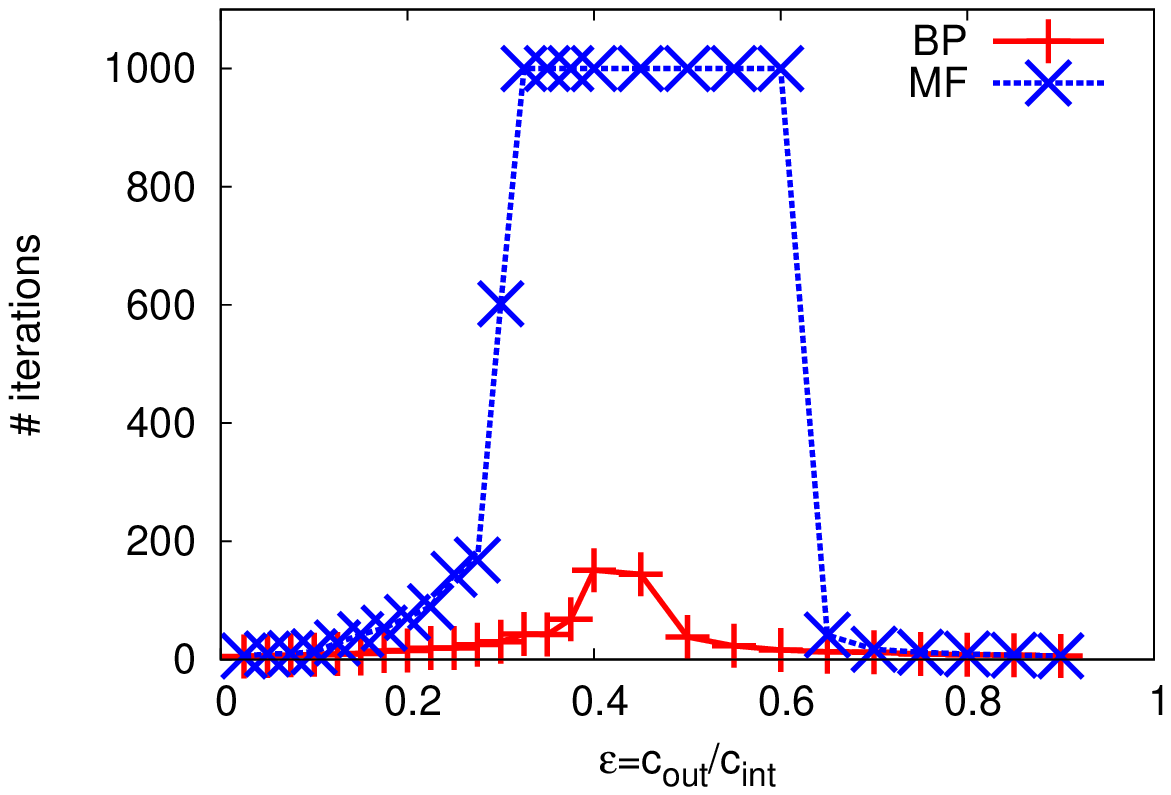}
 \includegraphics[width=0.32\linewidth]{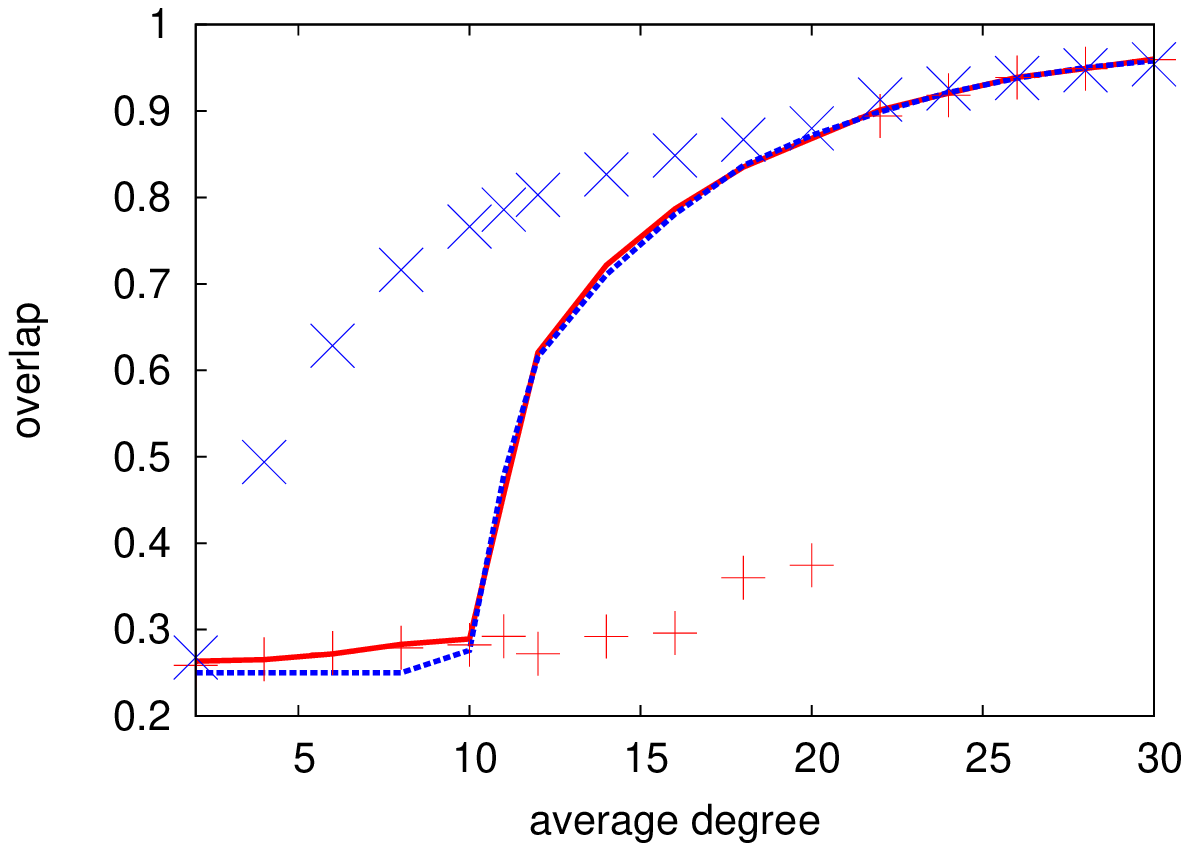}
  \caption{Comparison between the na\"{\i}ve mean field (MF) and belief propagation (BP) approaches to the E-step of expectation maximization. All datapoints correspond to networks with $N=10^4$ nodes. 
The networks were generated using $q=4$ groups, modular structure as sketched in left part of Fig.~\ref{fig_bloc}, and $c_{rr}=c_{\rm in}>c_{rs}=c_{\rm out}\, \forall s\neq r$.
  \textbf{Left:} True and illusive overlap $Q$ and $C$ for inference of the group assignment at different values of $\epsilon=c_{\rm out}/c_{\rm in}$. Note the transition between a phase where inference of class membership is possible and where it is not at $\epsilon_c=0.43$. Also note that MF is overfitting the data, showing large illusive overlap $C$ in the region where inference is in fact impossible. 
  \textbf{Middle:} The number of iterations needed for convergence of the E-step for the problem instances from the left part (we set the maximum number of iterations to be 1000). The computational effort is maximal at around $\epsilon_c$ for both methods, but BP converges faster. 
  \textbf{Right:} True and illusive overlap $Q$ and $C$ at different values of the average connectivity $c=\langle k\rangle$ at fixed $\epsilon=0.35$. Again, we observe a transition between feasible and infeasible inference at $\langle k\rangle_c(\epsilon)$ and the over-confidence of MF in the infeasible region. 
}
\label{fig1}
\end{figure*}

To compare the performance of BP and MF, we generated networks from the ``four groups test'' of \cite{PhysRevE.69.026113} with a large number of variables $N$, four groups $q=4$, average degree $c=p_0/N$, and ratio $\epsilon$ between the probability of being connected to a different group and within the same group. In other words, $\epsilon=c_{\rm out}/c_{\rm in}$. See an example adjacency matrix in Fig.~\ref{fig_bloc}.  The results of inference using BP and MF are plotted in Fig.~\ref{fig1}. 
From Fig.~\ref{fig1} we see several important points in which BP is superior over MF
\begin{itemize}
   \item BP estimate gives better agreement with the true assignment. In the left and right part of Fig.~\ref{fig1} we see the following. In a region of large overlap, the two methods give the same result. This can be understood from the form of the BP and MF equations that become equivalent for very polarized marginals $\psi^i_r$. In the region of very small overlap both approaches converge to a fixed point that does not contain any information about the original group assignment.  However, for parameter values close to the possible-impossible-inference phase transition the BP method gives larger overlap with the true group assignment than MF.
   \item BP is not over-confident. In the left and right part of Fig.~\ref{fig1} we compare the true overlap to the confidence value (\ref{over_conf}). For BP the two agree, just as they should if the marginals were evaluated exactly. In the MF approach, however, the confidence is considerably larger than the true overlap. This means that in the whole region where $C-Q > 0$, MF is misleadingly confident about the quality of the fixed point it found. The width of this region depends on the parameter values, but we observed that a good rule of thumb is that if the overlap reached is not very close to 1, then the MF method is unreliable. 
   \item BP is faster. As we explained when we exposed the BP and MF equations, one iteration takes a comparable time for both methods. In the middle part of Fig.~\ref{fig1} we plot the number of iterations needed for convergence, we see that again around the phase transitions region MF needs more iterations to converge, and hence is overall slower that BP. 
   \item BP does not converge to several different fixed points. Starting with randomly initialized messages, BP converged to the same fixed point (up to small fluctuations) in all the runs we observed. On the other hand in the region where the MF value of confidence $C$ differs from the true overlap $Q$ MF converged to several different fixed points depending on the initial conditions. 
\end{itemize}
To summarize, BP for block model inference is superior to MF in terms of speed, of quality of the result and does not suffer from over-confidence the way MF does. Note that similar conclusions about BP compared to MF were reached for other inference problems in e.g. \cite{Weiss_inbook_2001,Getoor}.

An important point is that so far, have have discussed the situation of BP and MF algorithms using the known and correct values of parameters $p_r$, $p_{rs}$ in the E-step of expectation maximization. Concerning the M-step, we observed without surprise that the expectation maximization with BP gives better results than with MF in the region of parameters where BP is superior for the E-step. Otherwise the performance was comparable. Notably, both the approaches suffer from a strong dependence on the initial conditions of the parameters $p_{rs}^{t=0}$. This is a known problem in general expectation maximization algorithms \cite{Karlis_Xekalaki_2003}. The problem comes from the fact that the log-likelihood $\mathcal{L}(\theta)$ has many local maxima (each corresponding to a fixed point) in $\theta$ in which the expectation maximization update gets stuck. Fortunately the free energy serves as an indicator of which fixed point of EM is better. Hence a solution is to run the EM algorithm from many different initial conditions and to consider the fixed point with the smallest free energy (i.e. largest likelihood). Since the volume of possible parameters does not grow in the system size $N$, this still leads to an algorithm linear in the system size (for sparse networks). However, the increase in the running time is considerable and smarter initializations of the parameters $p_{rs}^{t=0}$ are desired. We introduce one such in the next section.

\subsection{Spectral methods}

 \begin{figure*}[!ht]
 \center{
    \includegraphics[width=0.4\linewidth]{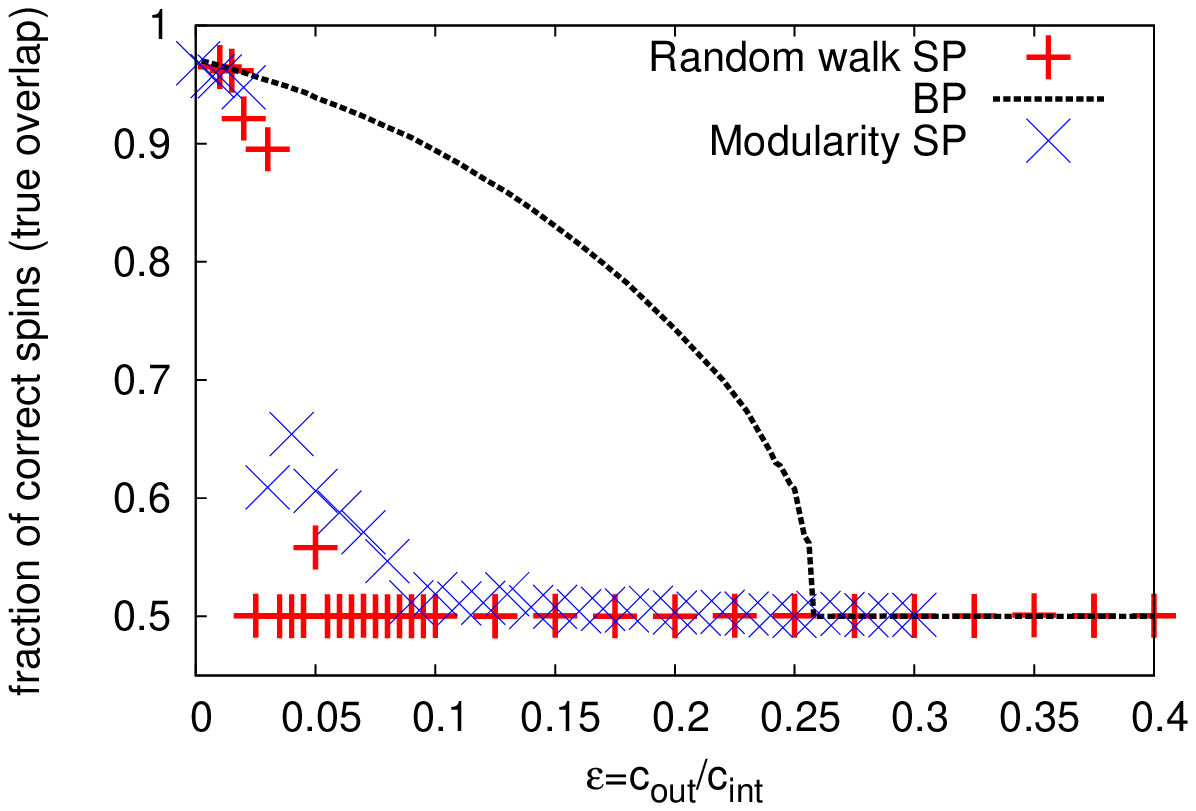}
    \includegraphics[width=0.4\linewidth]{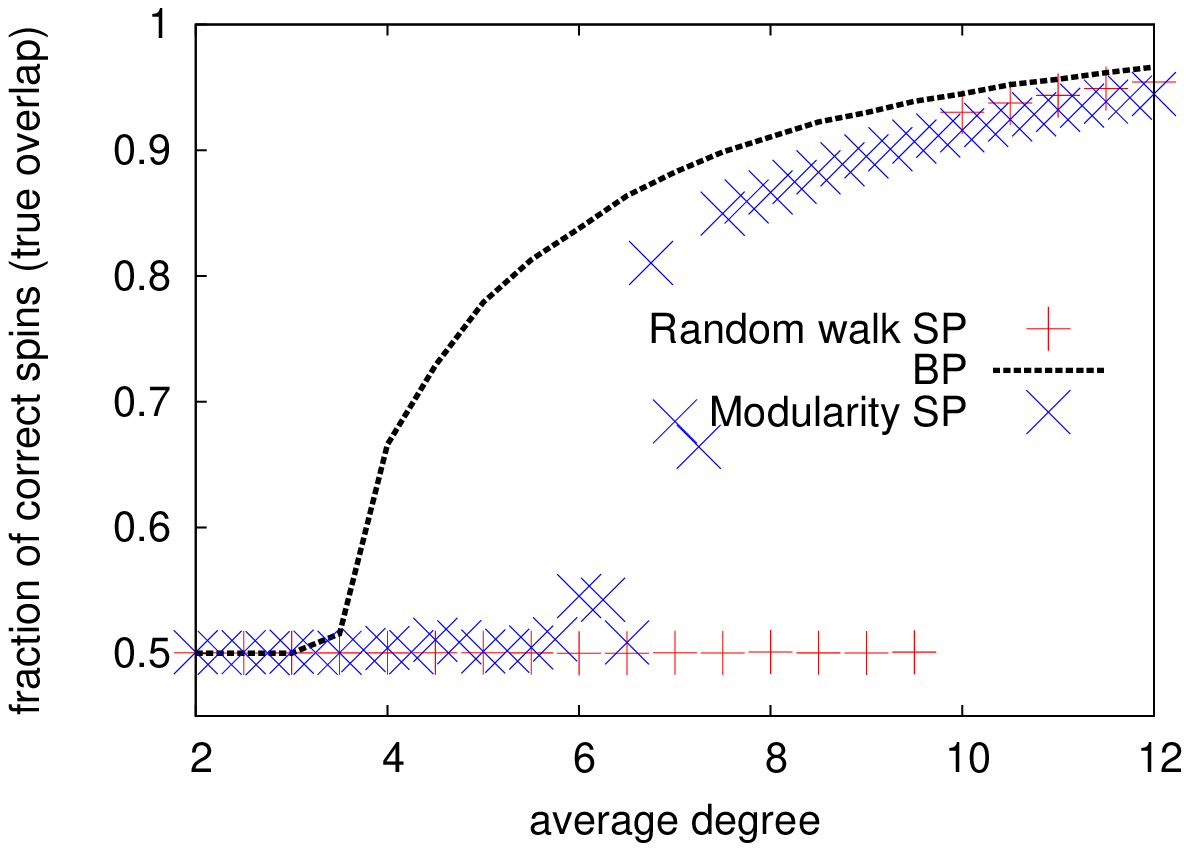}
}
    \caption{ Comparison of the BP (only the E-step) and spectral clustering based on the random walker approach and the modularity matrix (see text). Datapoints correspond to networks with $N=10^6$ nodes for the two spectral approaches and $N=10^5$ for BP, and average degree $c=\langle k \rangle=3$. The networks of $q=2$ groups are generated using modular structure as sketched in the left part of Fig.~\ref{fig_bloc}. To ensure the random walk based method to work, we extracted the largest connected component of the network and ran the algorithm on it. 
    \textbf{Left: } The overlap $Q$ at different values of $\epsilon=c_{\rm out}/c_{\rm in}$. Note how the spectral approaches can only correctly recover the latent class labels deep in the feasible region of the parameter space. 
    \textbf{Right: } The overlap $Q$ at different values of the connectivity $c$ at fixed $\epsilon=0.3$. Again, the spectral methods can only identify the latent class labels for problem instances well within the feasible region and fail on the hard instances near the critical connectivity. 
    }
    \label{fig2}
\end{figure*}

\begin{figure*}[!ht]
 \center{
 \includegraphics[width=0.4\linewidth]{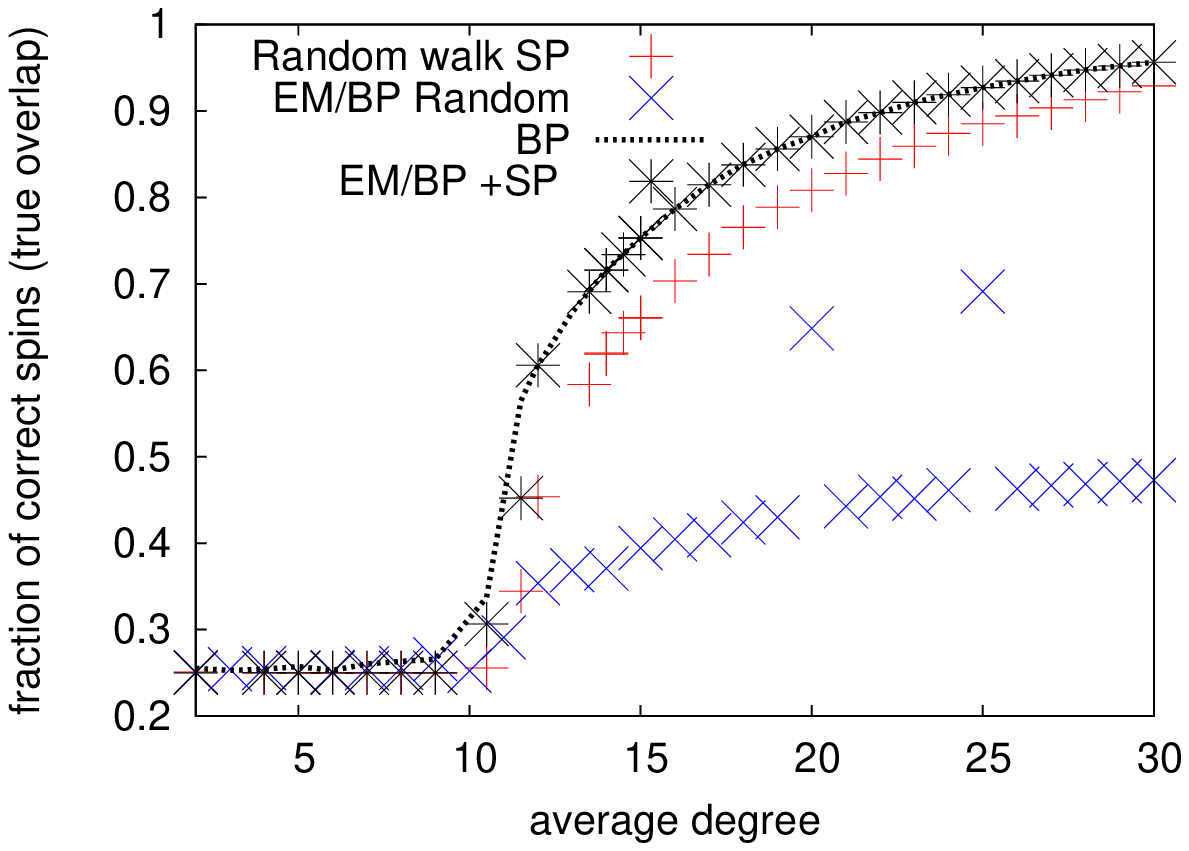}
\includegraphics[width=0.4\linewidth]{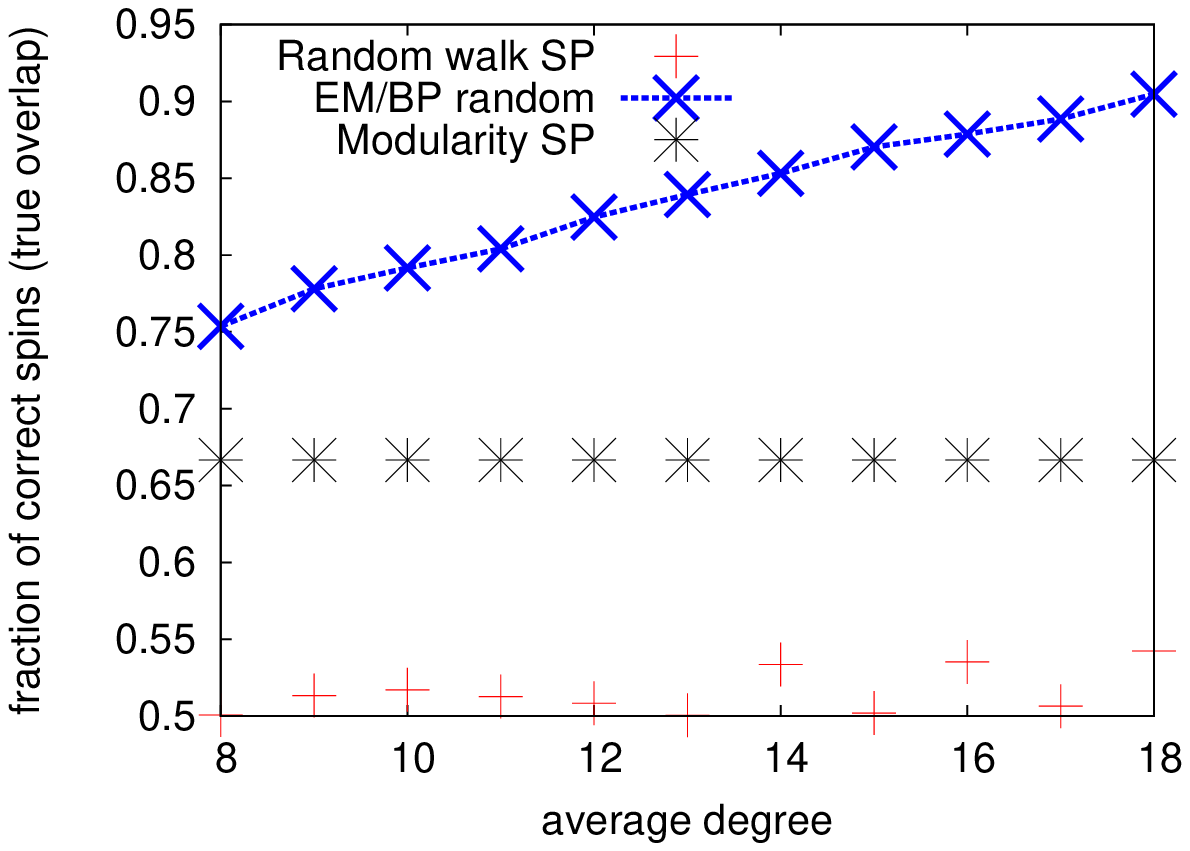}
}
\caption{   
\textbf{Left:} An example where the EM with BP when initialized in a random matrix $c_{ab}$ does not work, whereas the random walker spectral method works well. The result of the spectral method serves as a initialization of $c_{ab}$ in the EM BP, which then improves the achieved overlap. Modular network of size $N=10^5$ generated with $q=4$ groups and $\epsilon=0.35$. 
\textbf{Right:} An example where EM with BP works well even from random initial condition for the matrix $c_{ab}$, while spectral methods do not work well at all. The network exhibits a core periphery structure (middle panel of Fig.\ref{fig_bloc}) of size $N=10^4$. Here average degree of core variables is equal to average degree of periphery variables.
  There are two groups of sizes $p_{a}=2/3$ and $p_{b}=1/3$, 
  $c_{ab}$ matrix is in form of \{$c_{\rm in},c_{io};c_{io},c_{\rm out}$\}, with $c_{\rm in}=\frac{9c}{8-\epsilon}$, $c_{\rm out}=\epsilon c_{\rm in}$ and $c_{io}=1-0.5\epsilon$.  The modularity based method gives overlap $2/3$, because all variables were assigned to group $1$.   }
\label{fig3}
\end{figure*}

Methods based on the eigenvectors of the adjacency matrix of the network provide one of the most flexible approaches of graph clustering problems applied in the practice and hence we compare the BP algorithm to this approach as well. The comparison of BP with modularity matrix based and random walker based spectral methods gives the following conclusions:

\begin{itemize}
\item In the case when the parameters $\theta$ are known and we search for the best estimate of the original group assignment we observed that BP is always better than the two spectral clustering algorithms (that is the random walker based and the modularity based one) that we tested. This is illustrated in Fig.~\ref{fig2} and \ref{fig3}. In some cases (e.g. Fig.~\ref{fig3} left) the improvement BP provides over spectral methods is marginal. In other cases, e.g. for the core-periphery network of Fig.~\ref{fig3} right the improvement is drastic.

\item A particularly important point we want to make is the following: For the cases tested in this paper the spectral methods are clearly suboptimal: there are regions where the BP inference gives large overlap while spectral clustering methods do not do better than chance. See for instance Fig.~\ref{fig2} left for $~0.1<\epsilon<0.268$. 
Recently authors of \cite{PhysRevLett.108.188701} claimed "No other method will succeed in the regime where the modularity method fails'', it was mentioned that their results may not be valid for networks with small average degree. Here we clearly show that for networks with small average degree the spectral methods are indeed not optimal. In our opinion, the conclusions of \cite{PhysRevLett.108.188701}  apply only when the average degree diverges with the system size. 

\item A final point is that the spectral method should thus not be thought as the end of the story, but rather as the beginning: Indeed, they are extremely useful as a starting point for initializing EM BP to achieve improved overlap. This is shown in Fig.~\ref{fig3} left where EM BP starts from parameters taken from the result of the random walker based spectral method. This clearly improves the quality of the inference without having to restart EM BP for many initial conditions. 
\end{itemize}

\section{Conclusions}
Using the example of latent variable inference in the stochastic block model of complex networks, we have compared belief propagation based inference techniques with traditional mean field approaches and classic spectral heuristics. To this end, we have used the recent discovery of a sharp transition in the parameter space of the stochastic block model  from a phase where inference is possible to a phase where inference is provably impossible. In the vicinity of the phase transition, we find particularly hard problem instances that allow a performance comparison in a very controlled environment.  

We could show that though spectral heuristics are appealing at first for their speed and uniqueness of the resulting decompositions, they only work reliably deep within the parameter region of feasible inference. In particular, very sparse graphs are difficult for spectral methods, as are block structures that are more complicated than a mere collection of cohesive subgraphs or communities.  In short, they serve as a ``quick and dirty'' approach. We also evaluate if recent claims on the optimality of  spectral methods for block structure detection hold for networks with small average degree \cite{PhysRevLett.108.188701}.

Comparing na\"{\i}ve mean field techniques with belief propagation techniques, we find that the computational burden, which has so far hindered the wide spread use of belief propagation in fully connected graphical models such as block structure inference of (sparse or dense) networks, has been lifted completely. Not only is the computational complexity of the variable updates the same, belief propagation also exhibits much better convergence properties and this in particular on the hard problem instances. Hence, we expect that the presented formulations of belief propagation equations may find a wide range of application also in other fields of inference with fully connected graphical models. Note that the regime of $p_{rs} = O(1/N)$ considered here corresponds to the maximally sparse case. BP will still outperform other methods when $p_{rs}=O(N^{-\alpha})$ with $\alpha<1$, albeit the performance differences will be much smaller. 

Finally, we could show that using spectral decompositions in order to select initial conditions for learning the parameters of the stochastic block model can be a viable step in order to reduce the dependency on initial conditions when used in conjunction with expectation maximization type algorithms.

\section*{Acknowledgments}
We wish to thank to Cris Moore for discussions about various aspects of the EM BP algorithm. This work was supported by the Projet DIM "probl\'ematique transversales aux syst\`emes complexes" of the Institut des Syst\`emes Complexes, Paris \`Ile-de-France (ISC-PIF). J.R. was supported by a Fellowship Computational Sciences of the Volkswagen Foundation.


\end{document}